\documentclass{article} 
\PassOptionsToPackage{table}{xcolor}
\usepackage{iclr2027_conference,times}

\usepackage{amsmath,amsfonts,bm}

\def\eqref#1{equation~\ref{#1}}

\def\1{\bm{1}}

\DeclareMathAlphabet{\mathsfit}{\encodingdefault}{\sfdefault}{m}{sl}
\SetMathAlphabet{\mathsfit}{bold}{\encodingdefault}{\sfdefault}{bx}{n}

\usepackage{multirow}
\usepackage{xcolor}
\usepackage{amssymb}
\usepackage{pifont}
\usepackage{booktabs}
\usepackage{graphicx}
\usepackage{wrapfig}
\usepackage{float}
\usepackage{caption}
\usepackage{algorithm}
\usepackage{algorithmic}
\usepackage{url}
\usepackage{needspace}
\usepackage[hidelinks]{hyperref}

\newcommand{\gcheck}{\textcolor{green!60!black}{\ding{51}}}
\newcommand{\rcross}{\textcolor{red!85!black}{\ding{55}}}
\definecolor{bestfill}{RGB}{255,220,160}
\definecolor{secondfill}{RGB}{219,234,254}
\definecolor{projectpink}{RGB}{255,90,150}

\title{Beyond Temporal Smoothing: Spatial Energy\\
Budgets Stabilize One-Step Diffusion Editing}

\author{
Shengxiao Zhou$^{1}$ \quad Lei Luo$^{1,*}$ \quad Jian Yang$^{2}$\\
\normalfont $^{1}$PCA Lab, Nanjing University of Science and Technology \quad $^{2}$Nankai University\\
\normalfont \texttt{\{shengxiao.zhou, cslluo\}@njust.edu.cn}\quad
\texttt{csjyang@nankai.edu.cn}
}

\iclrfinalcopy

\begin{document}

\maketitle
\lhead{}

\begingroup
\centering
\vspace{-1.8em}
\includegraphics[width=\textwidth]{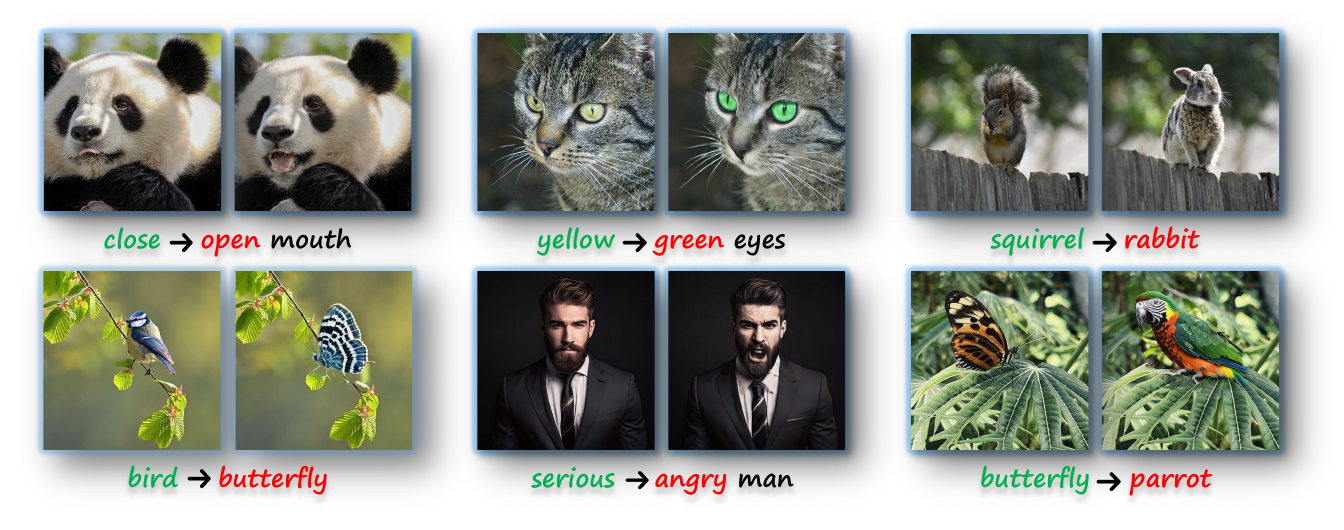}
\captionsetup{skip=4pt}
\captionof{figure}{\textbf{One-step editing by BudEdit.}
It converts background residual energy into a spatial budget and allocates the
constructed injection only to edit-relevant regions for stronger, more coherent edits.}
\label{fig:edits}
\endgroup
\vspace{1em}

\begin{abstract}
One-step text-guided diffusion editing is efficient but prone to spatially misallocated updates that distort the edited object and alter the background. Existing methods often improve stability by averaging the editing field across timesteps. We instead identify spatial energy misallocation as a distinct and measurable failure mode: across two independent noise draws, the residual field is essentially unrepeatable, making the background field unreliable for direct transport, while its total energy still sets a usable magnitude for the draw at hand. BudEdit turns that magnitude into an explicit budget and reallocates it to edit-relevant regions selected jointly by residual energy and cross-attention, controlling where editing energy is spent rather than averaging over timesteps. The resulting training-free, inversion-free editor spends the budget on transport and reuses it to scale a correction in a
lower-noise gated refinement. The budgeted injection field matches its prescribed budget exactly and vanishes on the identified background support, by construction. On PIE-Bench with SD-Turbo, BudEdit outperforms ChordEdit under each method's reported default settings on all 11 evaluated metrics, including a $2.1$\,dB gain in background PSNR, $31$\% lower DINO, and $36$\% lower LPIPS, while improving all five editing-quality metrics and reporting the lowest runtime in the comparison. \href{https://xiao0219.github.io/BudEdit/}{\textcolor{projectpink}{\textbf{Project}}}
\end{abstract}

\section{Introduction}
\label{sec:intro}

Distilled one-step generators such as
SD-Turbo~\citep{sauer2023adversarial},
InstaFlow~\citep{liu2023instaflow}, and SwiftBrush~\citep{nguyen2023swiftbrush}
have made real-time synthesis practical, but extending that speed to
text-guided editing has proven harder: the editing signal is the
difference between two independently conditioned model outputs (source and
target prompts), and
integrating this field in a single large step can produce object distortion
and background degradation. Existing training-free editors~\citep{kulikov2025flowedit,xu2024infedit}
stay stable by distributing the update over many steps, while trained
one-step editors~\citep{nguyen2025swiftedit} require task-specific training
and therefore target a different setting from training-free approaches.

\setlength{\intextsep}{0pt}
\needspace{16\baselineskip}
\begin{wrapfigure}{r}{0.5\textwidth}
\captionsetup{skip=1pt, belowskip=0pt}
\vspace{-3pt}
\centering
\includegraphics[width=\linewidth]{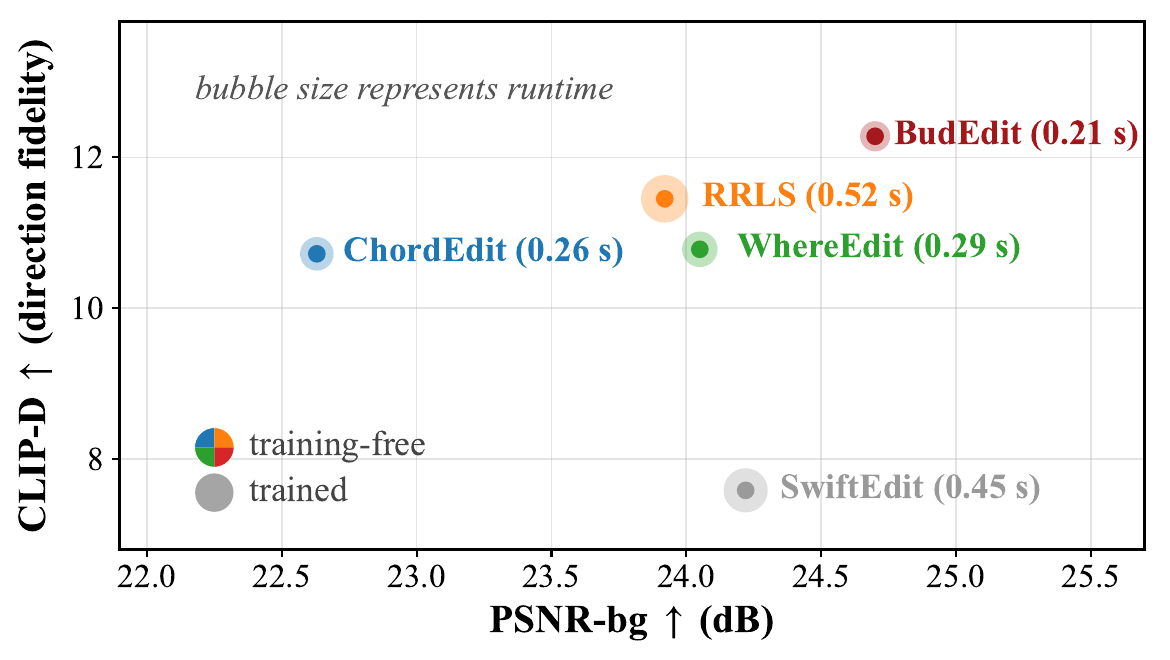}
\caption{\textbf{One-step editors on PIE-Bench.} PSNR-bg vs.\
CLIP-D, bubble size = runtime; BudEdit leads both axes and reports the
lowest runtime in this comparison.}
\label{fig:frontier}
\end{wrapfigure}

Recent one-step editors have explored different approaches to stabilizing this regime. ChordEdit~\citep{lu2026chordedit} attributes the instability to the volatility of the single-timestep field and proposes temporal smoothing as a remedy. Other recent methods infer an edit mask and amplify conditional transport,
as in WhereEdit~\citep{whereedit}, or refine the transport trajectory with
the Riemannian residual line search of RRLS~\citep{rrls}, neither of which
controls where the single-draw field deposits its energy. This motivates a
complementary question: can the spatial distribution of the editing field be
controlled explicitly rather than averaged uniformly across timesteps?

We identify spatial energy misallocation as a measurable failure mode of
one-step editing. Across two independent noise draws the residual field is essentially unrepeatable, with region-level cosine similarity near $0.1$ in both the injection support and its complement, and its amplitude map is no more stable. 
The source-anchored displacement is substantially more reproducible and is therefore used as the injection direction. Although the residual remains draw-dependent, BudEdit uses its realized energy as a budget scale for the current draw and its high-energy support as a set of candidate edit locations for that draw. Accordingly, the raw residual is not transported globally; it is retained only as an attenuated, support-restricted base term.
The appendix reports these
two-draw measurements.

We therefore introduce BudEdit, a training-free and inversion-free editor that
converts unreliable background residual energy into an image-dependent budget
and reallocates it to edit-relevant regions. A residual-energy support and a
cross-attention localization signal, both read off the transport forward pass at
no additional UNet evaluation, determine where the budget is allocated,
while a constrained allocation problem determines how it is distributed.
The resulting closed-form construction provides exact budget matching on
nonempty injection support and hard-support control for the injected field. A second low-noise pass then cleans
up the transported image: within an attention-derived gate, GatedRefine scales its
correction by the same budget and restores source detail.

Our contributions are summarized as follows:
\begingroup\setlength{\leftmargini}{0pt}\setlength{\itemsep}{0pt}\begin{itemize}
  \item 
  We identify spatial energy misallocation as a measurable failure mode in one-step editing, complementing existing views centered on temporal instability.

  \item
  We formulate spatial energy budgeting as a constrained allocation problem and derive a closed-form solution with exact construction-level budget matching on nonempty injection support and hard-support control.

  \item 
  We propose BudEdit, which sets its budget from background residual energy and injects it where the energy support meets a prompt-difference attention gate, outperforming the temporal-smoothing baseline on all 11 PIE-Bench metrics.
\end{itemize}\endgroup

\section{Related Work}
\label{sec:related}

Multi-step editing established the prevailing recipe for stability: shape
how the edit signal enters the denoiser and pay for the control with
steps, typically 20--50 evaluations per edit. Attention manipulation and
spatial feature injection guide the edit from inside the model~\citep{hertz2022prompt,tumanyan2023plug},
while inversion-based pipelines reconstruct the source along a guided
trajectory~\citep{song2020denoising,mokady2023nulltext}.
Distilled generators~\citep{salimans2022distillation,song2023consistency,sauer2023adversarial,liu2023instaflow,nguyen2023swiftbrush}
offer real-time synthesis, but their editing signal, a difference of two
independently conditioned predictions, must be integrated in a single large
step; existing inversion-free editors~\citep{kulikov2025flowedit,xu2024infedit}
stay above interactive rates, and FlowDC~\citep{jiang2025flowdc} likewise
operates over multiple steps while decaying the velocity components
orthogonal to the edit direction for multi-target complex editing on its own
Complex-PIE-Bench; trained
editors~\citep{brooks2023instructpix2pix,nguyen2025swiftedit} require
task-specific training rather than the training-free operation studied here.

Among training-free one-step editors, ChordEdit~\citep{lu2026chordedit}
sets the reference: it frames the task as dynamic optimal transport,
attributes instability to the volatility of the single-timestep field,
and stabilizes integration by averaging observable residuals at two
timesteps. WhereEdit~\citep{whereedit} performs mask-aware local latent
editing and RRLS~\citep{rrls} refines the transport with a Riemannian
residual line search. These methods leave open the question of whether
measured residual energy can be controlled explicitly in space.

Across these approaches, stabilization remains implicit. To our knowledge, no
training-free one-step editor formulates the spatial allocation of measured
residual energy as an image-dependent budget with construction-level support
constraints.

Our work makes this energy explicit. In the Benamou--Brenier
formulation~\citep{benamou2000computational}, kinetic energy is the
fundamental transport cost, and low-energy paths underlie optimal-transport
flow matching~\citep{lipman2022flowmatching}; editing, however, has not
treated measured residual energy as a controllable quantity. We do: the
single-draw editing field is read as a noisy measurement, its draw-sensitive
background energy is converted into an image-dependent budget, and the budget
is re-injected where the model's own attention indicates edit relevance.
This reframes stabilization from uniform temporal smoothing to explicit
spatial allocation of editing energy.

\section{Method}
\label{sec:method}

\subsection{Preliminaries}
\label{sec:prelim}

Following \citet{lu2026chordedit}, we use the probability-flow ODE
framework~\citep{song2021score}. A text-to-image model conditioned on $c$ induces
a drift $v(x_t,t,c)$; editing transports the source latent $x_{\mathrm{src}}$ from the source-conditioned data distribution $p(x\,|\,c_{\mathrm{src}})$ toward the target-conditioned one $p(x\,|\,c_{\mathrm{tar}})$. At current latent
$x_\tau$, which is $x_{\mathrm{src}}$ in the single-step setting, and time $t$,
the model is queried at $z\sim K_t(\cdot\,|\,x_\tau)$. Its
output $Q(z,t,c)$ (noise prediction for SD-Turbo, velocity for flow models) is
mapped by the time-only linear map $\mathcal{T}_t$ into a common comparison field,
and the source/target predictions share the same $z$.

At fixed $(x_\tau,t)$, write $Q_z^c \equiv Q(z,t,c)$. The algorithm computes a single-draw residual, while its population counterpart averages over the noise distribution:
\begin{equation}
\widehat{R}_z
\equiv
\widehat{R}(x_\tau,t;z)
=
\mathcal{T}_t\!\left(
Q_z^{c_{\mathrm{tar}}}
-
Q_z^{c_{\mathrm{src}}}
\right),
\qquad
\overline{R}
\equiv
\overline{R}(x_\tau,t)
=
\mathbb{E}_{z\sim K_t(\cdot\mid x_\tau)}
\left[\widehat{R}_z\right].
\label{eq:population_residual}
\end{equation}

We define the population residual field and its draw-dependent deviation as
\begin{equation}
u_t(x_\tau)
:=
\overline{R}(x_\tau,t),
\qquad
\epsilon_t(z)
:=
\widehat{R}_z-u_t(x_\tau).
\label{eq:residual_decomposition}
\end{equation}
By construction, $\mathbb{E}_{z\sim K_t(\cdot\mid x_\tau)}\left[\epsilon_t(z)\right]=0.$
Thus, $\widehat{R}_z=u_t(x_\tau)+\epsilon_t(z)$
decomposes the single-draw residual into its population mean and
draw-dependent deviation. When the noise draw is fixed, we suppress $z$
and write $\widehat{R}(x_\tau,t)$.

Throughout, $\|\cdot\|_C$ denotes the per-pixel channel-wise norm of a
latent field, and 
$\|h\|^2
\equiv
\sum_{(i,j)}\|h(i,j)\|_C^2$ 
denotes its squared $L_2$ norm over the domain $\Omega$, corresponding to
the entire image. We refer to this quantity as its total energy.

\subsection{An Energy Diagnosis of One-Step Instability}
\label{sec:diagnosis}

For the time-dependent ODE $\dot{x}=v(x,t)\equiv v_t(x)$, assume that $v$ is
$C^{2}$ with bounded derivatives along the exact trajectory and that the
Euler and exact solutions start from the same state. The local truncation
error of one explicit Euler step of size $\Delta t$ is bounded by the total
derivative of the vector field:
\begin{equation}
  \|x_{\mathrm{true}}-x_{\mathrm{euler}}\|
  \leq \frac{(\Delta t)^2}{2}
  \sup_{s\in[t,t+\Delta t]}
  \left\|\partial_s v(x(s),s)
  +D_xv(x(s),s)v(x(s),s)\right\| .
  \label{eq:gronwall}
\end{equation}
A useful upper bound replaces the total
derivative norm by
$\sup_s\|\partial_s v(x(s),s)\|+
\sup_s\|D_xv(x(s),s)\|\sup_s\|v(x(s),s)\|$. Thus, field magnitude is one
component of the one-step error. On the background support this argues for suppression, since a zero field removes its magnitude-dependent contribution. Inside the injection support the construction raises field magnitude instead, trading a larger local discretization error for a semantic transition completed in a single step.

For the conceptual diagnosis, let $\mathcal{E}^{\star}$ denote the
semantic edit region and let $\mathcal{B}^{\star}=\Omega\setminus\mathcal{E}^{\star}$
denote its background. We write the pointwise decomposition
$\widehat{R}(x_\tau,t;z)=u_t(x_\tau)+\epsilon_t(z)$. On
$\mathcal{B}^{\star}$, the desired semantic transport component underlying
$u_t$ is expected to be small, but the full population residual need not
vanish: it may contain systematic prompt-pair response as well as
draw-sensitive variation. We treat noise dominance as an operational
property, verified empirically rather than derived from the model: raw residual
amplitudes are draw-sensitive, and the background residual field is unreliable for direct transport. The corresponding two-draw measurement is reported in the Appendix.
For $0<\delta<t_s$, ChordEdit's chord control field averages residuals across
the two timesteps $t_s$ and $t_s-\delta$:
\begin{equation}
  \widehat{u}_{t_s}(x_\tau)
  = \frac{t_s\,\widehat{R}(x_\tau,t_s-\delta)
        + \delta\,\widehat{R}(x_\tau,t_s)}{t_s+\delta}.
  \label{eq:chord}
\end{equation}
The average is applied uniformly across edited and background regions, so it
reduces the variance of the field without deciding where its energy lands.

\begin{figure}[!t]
\centering
\includegraphics[width=\textwidth]{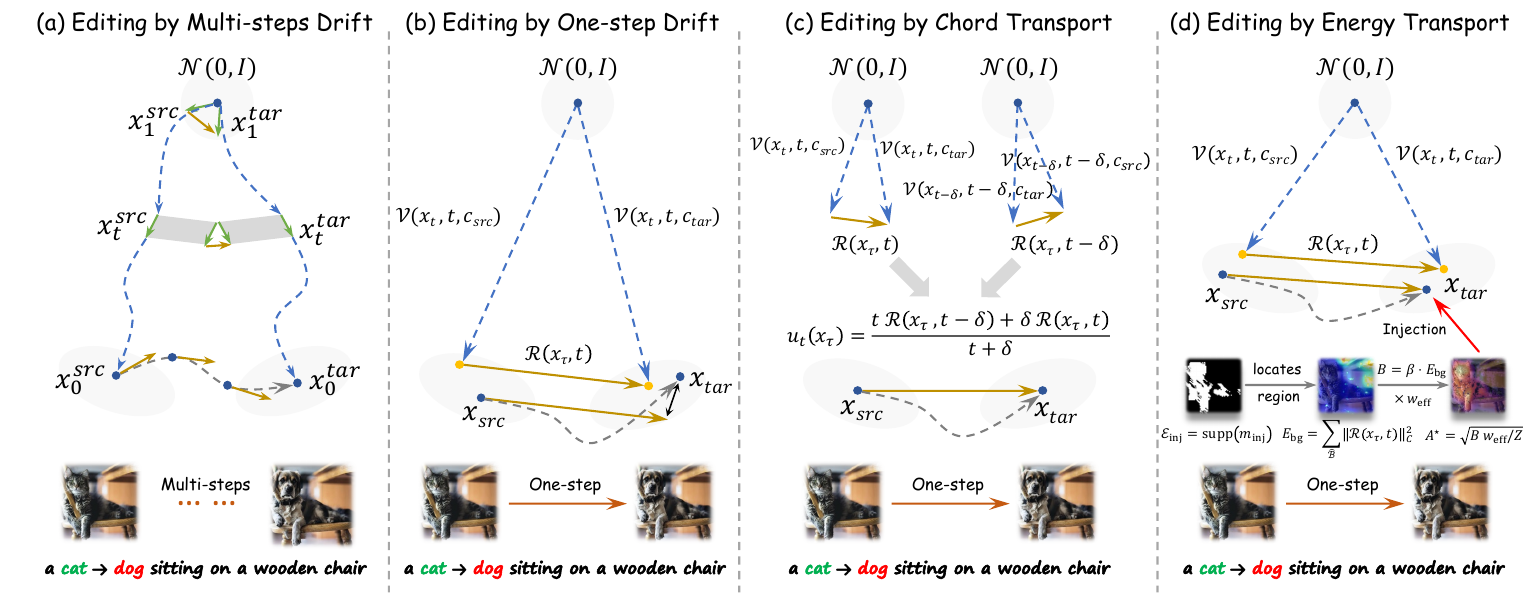}
\caption{\textbf{Comparison of editing paradigms.} (a) multi-step editing buys stability by spreading the update over many small steps; (b) one-step editing integrates the measured field in a single large step; (c) chord transport forms the one-step update from a time-weighted average of residual fields at two timesteps; (d) energy transport converts background residual energy into a budget and spends it inside the effective injection support.}
\label{fig:paradigms}
\end{figure}

Accordingly, BudEdit separates how much editing energy is administered from where and how it is spent: the spatially aggregated background residual energy from the current draw sets the per-image budget scale, while the effective injection support and allocation profile determine its spatial
deployment. The next section formalizes this allocation as a constrained optimization with a closed-form solution. Figure~\ref{fig:paradigms} contrasts this spatial budgeting view with temporal smoothing.

\subsection{Spatial Energy Budgeting}
\label{sec:principle}

The analysis above suggests a two-stage construction: the aggregate residual energy measured on the operational background support scales an image-dependent budget, which is then allocated within an effective edit support according to an explicit spatial profile.
\paragraph{Harvesting the budget.}
The algorithm estimates a data-dependent energy support
$\widehat{\mathcal E}$ from the residual field: it thresholds the per-pixel
residual energy and then applies a narrow boundary extension, represented by
a soft mask $m\in[0,1]^{H\times W}$ whose nonzero support is $\operatorname{supp}(m)=\widehat{\mathcal E}$. Let
$\widehat{\mathcal B}=\Omega\setminus\widehat{\mathcal E}$ denote
the corresponding operational background support.

Let $\Phi\in[0,1]^{H\times W}$ denote the peak-normalized cross-attention localization
signal and $\widehat{x}_{\mathrm{tar}}$ the target-conditioned latent
prediction, both read off the same UNet forward pass; the construction of
$\Phi$ is detailed in Section~\ref{sec:attention}. The residual-energy support
$\widehat{\mathcal E}$ provides candidate edit locations, while $\Phi$ gates and
weights them. With $\mathbf{1}[\cdot]$ denoting an indicator, we define the mask
pointwise for $(i,j)\in\Omega$ and set its active support as follows:
\begin{equation}
m_{\mathrm{inj}}(i,j)
=m(i,j)\,\mathbf{1}[\Phi(i,j)>0.5]
\,\mathbf{1}\!\left[\|\widehat{x}_{\mathrm{tar}}(i,j)-x_{\mathrm{src}}(i,j)\|_C>0\right],\quad
\mathcal E_{\mathrm{inj}}=\operatorname{supp}(m_{\mathrm{inj}}).
\label{eq:injsupport}
\end{equation}

The smoothed attention readout
$w=\operatorname{blur}(\Phi)$ specifies the within-support allocation
profile. Together with the soft injection mask, it gives
\begin{equation}
w_{\mathrm{eff}}(i,j)
=
m_{\mathrm{inj}}(i,j)^2\,w(i,j).
\label{eq:weight}
\end{equation}

Because $m_{\mathrm{inj}}$ scales field amplitude, its square enters the energy
profile. The nonnegative smoother satisfies $w(i,j)>0$ wherever
$\Phi(i,j)>0.5$, so Eq.~\ref{eq:injsupport} gives $w_{\mathrm{eff}}>0$ throughout
$\mathcal E_{\mathrm{inj}}$.

Energy is harvested from the operational background support, without applying
the attention gate:
\begin{equation}
E_{\mathrm{bg}}
=
\sum_{(i,j)\in\widehat{\mathcal B}}
\|\widehat{R}(i,j)\|_C^2,
\qquad
B=\beta E_{\mathrm{bg}}.
\label{eq:budgetdef}
\end{equation}
For $\beta\geq0$, $B\geq0$. The quantity $E_{\mathrm{bg}}$ sets an
image-dependent scale rather than a conserved quantity: $\beta=1$ equates the
budget with the measured background energy, while $\beta>1$ favors stronger
editing. We use $\beta=4.0$ and report the sweep in the Appendix. On nonempty support, Eq.~\ref{eq:budgetdef} makes the mean injected energy
density scale with $|\widehat{\mathcal B}|/|\mathcal E_{\mathrm{inj}}|$, giving
a small target on a large canvas a denser correction.

\paragraph{Allocation as information projection.}
Assume that $\mathcal{E}_{\mathrm{inj}}$ is nonempty and $B>0$.
Eq.~\ref{eq:injsupport} makes the displacement nonzero on this support, so
\begingroup
\setlength{\abovedisplayskip}{10pt}
\setlength{\belowdisplayskip}{10pt}
\begin{equation}
d(i,j)=\frac{\widehat{x}_{\mathrm{tar}}(i,j)-x_{\mathrm{src}}(i,j)}
{\|\widehat{x}_{\mathrm{tar}}(i,j)-x_{\mathrm{src}}(i,j)\|_C},\quad
\|d(i,j)\|_C=1,\quad (i,j)\in\mathcal E_{\mathrm{inj}}.
\label{eq:direction}
\end{equation}
\endgroup
The injection field takes the form $f=A\odot d$, where $A\ge0$ is the unknown
scalar amplitude to be solved for. We further define, for $f\not\equiv0$ on $\mathcal E_{\mathrm{inj}}$,
\begin{equation}
\mu_{w_{\mathrm{eff}}}(i,j)
=
\frac{w_{\mathrm{eff}}(i,j)}
{\sum_{(k,l)\in\mathcal{E}_{\mathrm{inj}}}w_{\mathrm{eff}}(k,l)},
\qquad
p_f(i,j)
=
\frac{\|f(i,j)\|_C^2}
{\sum_{(k,l)\in\mathcal{E}_{\mathrm{inj}}}\|f(k,l)\|_C^2}.
\label{eq:measures}
\end{equation}

For $B>0$, the budgeted solution is
\begin{equation} 
  (f^{\star},A^{\star})
  \;\in\; \arg\min_{f,A}\; 
  D_{\mathrm{KL}}\!\big(p_f \,\big\|\, \mu_{w_{\mathrm{eff}}}\big) 
  \;\text{s.t.}\; 
  \begin{aligned} 
    &\operatorname{supp}(f)\subseteq\mathcal{E}_{\mathrm{inj}},\qquad 
    f|_{\mathcal{E}_{\mathrm{inj}}} = A\odot d,\\[1pt] 
    &A(i,j)\geq 0,\qquad 
    \textstyle\sum_{(i,j)\in\mathcal{E}_{\mathrm{inj}}} A(i,j)^2 
    \;=\; B, 
  \end{aligned} 
  \label{eq:program} 
\end{equation}

The distributions induced by feasible solutions of Eq.~\ref{eq:program} include $\mu_{w_{\mathrm{eff}}}$, so the minimum is attained exactly at $p_{f^{\star}}=\mu_{w_{\mathrm{eff}}}$. With this and the fixed-budget constraint, the nonnegativity condition yields the unique amplitude map on $\mathcal E_{\mathrm{inj}}$:
\begin{equation}
A^\star(i,j)
=
\sqrt{
\frac{B\,w_{\mathrm{eff}}(i,j)}
{\sum_{(k,l)\in\mathcal{E}_{\mathrm{inj}}}w_{\mathrm{eff}}(k,l)}
},
\qquad
f^\star=A^\star\odot d.
\label{eq:closedform}
\end{equation}

For $B=0$ or $\mathcal{E}_{\mathrm{inj}}=\varnothing$, we set
$f^\star=0$; the implementation uses the same fallback if the normalizing sum
vanishes, a defensive case excluded by the positivity above. The nominal
transport field is
\begin{equation}
F
=
\alpha\,m_{\mathrm{inj}}\odot\widehat{R} + f^\star,
\qquad
x_{\mathrm{edit}}
= x_{\mathrm{src}}+F,
\label{eq:field}
\end{equation}
where $\alpha=0.7$. All image-dependent quantities in
Eqs.~\ref{eq:injsupport}--\ref{eq:field} are computed from the current image and prompts;
$\alpha$, $\beta$, and other implementation constants are fixed globally,
while the Appendix summarizes the remaining construction rules.

\paragraph{Exact injection-budget matching and background support.}
Under the assumptions above, the unique solution $f^{\star}$ of
Eq.~\ref{eq:program} satisfies
\begin{equation}
  \sum_{(i,j)\in\mathcal{E}_{\mathrm{inj}}}
  \|f^{\star}(i,j)\|_C^2 = B,
  \qquad
  f^{\star}|_{\widehat{\mathcal B}}=0 .
  \label{eq:budgetmatch}
\end{equation}

For $B>0$, Eq.~\ref{eq:budgetmatch} holds exactly; for $B=0$, both
sides are zero by convention. These guarantees apply to the administered
injection field $f^\star$. The full field $F$ also contains the attenuated
residual term, while refinement and fusion determine the final image-space
transition.

\paragraph{A transport-field bound on the injection support.}
By the triangle inequality and $0\leq m_{\mathrm{inj}}\leq1$, the composed field $F$
of Eq.~\ref{eq:field} satisfies:
\begin{equation}
  \sum_{(i,j)\in\mathcal{E}_{\mathrm{inj}}}\|F(i,j)\|_C^2
  \;\le\;\left(\alpha\sqrt{\sum_{(i,j)\in\mathcal{E}_{\mathrm{inj}}}
  \|\widehat{R}(i,j)\|_C^2}+\sqrt{B}\right)^{2}.
  \label{eq:budget}
\end{equation}
The bound controls the nominal composed field; the ODE discretization term
is governed by the local-error analysis above, while fixed calibration,
refinement, and output blending determine the final image-space transition.

\subsection{Localization from Prompt-Difference Attention}
\label{sec:attention}
\begin{wrapfigure}{r}{0.5\textwidth}
\captionsetup{skip=1pt, belowskip=0pt}
\vspace{0pt}
\begin{minipage}{\linewidth}
\centering
\footnotesize
\hrule height 1.2pt
\vspace{1pt}
\begingroup
\makeatletter
\def\@captype{algorithm}
\makeatother
\caption{BudEdit: energy transport.}\label{alg:budedit}
\endgroup
\vspace{1pt}
\hrule
\vspace{2pt}
\begin{algorithmic}[1]
\STATE \textbf{Input:} $x_{\mathrm{src}}$, $c_{\mathrm{src}},c_{\mathrm{tar}}$, $t_s,t_e,\alpha,\beta$
\STATE $z\gets\operatorname{noisify}(x_{\mathrm{src}},t_s)$
\STATE $(Q_z^{c_{\mathrm{src}}},Q_z^{c_{\mathrm{tar}}},\widehat{x}_{\mathrm{tar}},\Phi)\gets\operatorname{UNet}(z,t_s;\,\cdot\,)$
\STATE $\widehat{R}\gets\mathcal{T}_{t_s}(Q_z^{c_{\mathrm{tar}}}-Q_z^{c_{\mathrm{src}}})$
\STATE $\widehat{\mathcal E}\!\gets\!\operatorname{Extend}(\operatorname{EnergySupport}(\widehat{R}))$; $\widehat{\mathcal B}\!\gets\!\Omega\setminus\widehat{\mathcal E}$
\STATE $m\gets\operatorname{Soften}(\widehat{\mathcal E})$; $w\gets\operatorname{blur}(\Phi)$
\STATE $(m_{\mathrm{inj}},\mathcal E_{\mathrm{inj}})\gets$ Eq.~\ref{eq:injsupport}; $w_{\mathrm{eff}}\gets m_{\mathrm{inj}}^2\odot w$
\STATE $Z\gets\sum_{(i,j)\in\mathcal E_{\mathrm{inj}}}w_{\mathrm{eff}}(i,j)$
\STATE $E_{\mathrm{bg}}\gets\sum_{(i,j)\in\widehat{\mathcal B}}\|\widehat{R}(i,j)\|_C^2$; $B\gets\beta E_{\mathrm{bg}}$
\STATE \textbf{if} $\mathcal E_{\mathrm{inj}}{=}\varnothing$ \textbf{or} $B{=}0$ \textbf{or} $Z{=}0$: $f^{\star}\!\gets\!0$
\STATE \textbf{else}: $d\!\gets\!\operatorname{unit}(\widehat{x}_{\mathrm{tar}}-x_{\mathrm{src}})$ on $\mathcal E_{\mathrm{inj}}$; $f^{\star}\!\gets\!A^{\star}\odot d$ by Eq.~\ref{eq:closedform}
\STATE $x_{\mathrm{edit}}\gets x_{\mathrm{src}}+\alpha m_{\mathrm{inj}}\odot\widehat{R}+f^{\star}$
\STATE $z_e\gets\operatorname{noisify}(x_{\mathrm{edit}},t_e)$
\STATE $(Q_{z_e}^{c_{\mathrm{src}}},Q_{z_e}^{c_{\mathrm{tar}}},\widetilde{x}_{\mathrm{tar},e},\Phi_{\mathrm{pin}})\gets\operatorname{UNet}(z_e,t_e;\,\cdot\,)$
\STATE $\widetilde{R}_e\gets\mathcal{T}_{t_e}(Q_{z_e}^{c_{\mathrm{tar}}}-Q_{z_e}^{c_{\mathrm{src}}})$
\STATE $(x_{\mathrm{refined}},\Psi)\gets\operatorname{GatedRefine}(\,\cdot\,)$ by Eq.~\ref{eq:gatedrefine}
\STATE $x_{\mathrm{out}}\gets\Psi\odot x_{\mathrm{refined}}+(1-\Psi)\odot x_{\mathrm{src}}$; \textbf{return} $x_{\mathrm{out}}$
\end{algorithmic}
\vspace{2pt}
\hrule height 1.2pt
\end{minipage}
\vspace{-4pt}
\end{wrapfigure}

Cross-attention maps bind spatial positions to text tokens~\citep{hertz2022prompt,chefer2023attend,whereedit}.
A capturing attention processor, swapped into the forward pass already
required for transport, yields an attention signal
$\Phi\in[0,1]^{H\times W}$ without an additional UNet evaluation: maps of the differing tokens are averaged over attention heads, upsampled to latent resolution, averaged over layers, peak-normalized, and finally max-combined per pixel across the two prompt sides. The blurred readout $w=\operatorname{blur}(\Phi)$ supplies the unweighted allocation signal. Together
with the soft injection mask, it forms the effective allocation weight
$w_{\mathrm{eff}}$ of Eq.~\ref{eq:weight}; the same measurement supplies the
pointwise hard attention gate of Eq.~\ref{eq:injsupport}.

Attention plays three roles: defining the hard attention gate, shaping the effective profile $w_{\mathrm{eff}}$ within the region, and gating the second-pass refinement of Section~\ref{sec:refinement}. Energy support supplies candidate edit locations; cross-attention gates and weights the injected field.

\subsection{Budget-Aware Gated Refinement}
\label{sec:refinement}

\paragraph{Second-pass cleanup.}
Given the transported prediction $x_{\mathrm{edit}}$, we introduce a second-pass
cleanup, implemented as a single forward pass at the lower noise level $t_e$ that
reuses the noise realization of the transport pass:
\begin{equation}
  z_e=\operatorname{noisify}(x_{\mathrm{edit}},t_e),
  \qquad
  \widetilde{R}_e
  =\mathcal{T}_{t_e}\big(Q_{z_e}^{c_{\mathrm{tar}}}-Q_{z_e}^{c_{\mathrm{src}}}\big).
  \label{eq:cleanup_pass}
\end{equation}
The same query also returns a target-conditioned prediction
$\widetilde{x}_{\mathrm{tar},e}$ and a fresh attention readout
$\Phi_{\mathrm{pin}}$. Thresholding $\Phi_{\mathrm{pin}}$, dilating the result,
and intersecting it with the energy support of $\widetilde{R}_e$ yields the
cleanup gate. Starting from $\widetilde{x}_{\mathrm{tar},e}$, within this gate we reinject
a fraction of $B$ as a residual correction, continuing the budgeted edit, and
apply a gated local extrapolation away from $x_{\mathrm{edit}}$ to recover
reliable but under-edited content.

\paragraph{Gated refinement and fusion.}
The cleanup candidate is further refined with a low-amplitude high-frequency source
anchor and an attention-modulated micro-swap, which restore local texture and
stabilize uncertain local structure. We write the refinement and its output gate as
\begin{equation}
  (x_{\mathrm{refined}},\Psi)
  =\operatorname{GatedRefine}\big(
    x_{\mathrm{edit}},x_{\mathrm{src}},m,
    \widetilde{x}_{\mathrm{tar},e},
    \widetilde{R}_e,\Phi_{\mathrm{pin}};\,B
  \big).
  \label{eq:gatedrefine}
\end{equation}
Here $\Psi\in[0,1]^{H\times W}$ retains the first-pass edit support where
second-pass attention admits it, and adds extensions jointly supported by the
second-pass residual and attention;
soft boundaries and a graded confidence ramp govern final fusion:
\begin{equation}
  x_{\mathrm{out}} = \Psi\odot x_{\mathrm{refined}}
                   + (1-\Psi)\odot x_{\mathrm{src}} .
  \label{eq:output}
\end{equation}

\section{Experiments}
\label{sec:experiments}

\subsection{Setup}
\label{sec:setup}

\paragraph{Dataset and baselines.}
PIE-Bench~\citep{ju2024pnp}: 700 images, 10 editing categories, all
$512\!\times\!512$, seed fixed at 42 for all methods. Baselines:
multi-step (SpotEdit~\citep{qin2025spotedit},
FlowEdit~\citep{kulikov2025flowedit},
KV-Edit~\citep{zhu2025kvedit},
PnPInversion~\citep{ju2024pnp}), few-step
(TurboEdit~\citep{deutch2024turboedit},
InstantEdit~\citep{instantedit},
InfEdit~\citep{xu2024infedit}), and one-step (trained
SwiftEdit~\citep{nguyen2025swiftedit}; training-free
ChordEdit~\citep{lu2026chordedit}, WhereEdit~\citep{whereedit},
RRLS~\citep{rrls}). All methods are evaluated using their respective reported default configurations.

\begin{table}[t]
\renewcommand{\arraystretch}{1.1}
\caption{\textbf{Quantitative comparison on PIE-Bench.} T-free: Training-free. I-free: Inversion-free. Best and second-best among one-step methods are highlighted in {\setlength{\fboxsep}{2pt}\colorbox{bestfill}{yellow}} and {\setlength{\fboxsep}{3pt}\colorbox{secondfill}{blue}}, ties included.}
\vskip -5pt
\label{tab:comparison}
\centering
\footnotesize
\setlength{\tabcolsep}{0.7pt}
\resizebox{\textwidth}{!}{
\begin{tabular}{cc|cccccc|ccccc|cc|cccc}
\toprule
\multicolumn{1}{c}{\multirow{2}{*}{\textbf{Type}}}
& \multicolumn{1}{c}{\multirow{2}{*}{\textbf{Methods}}}
& \multicolumn{6}{c}{\textbf{Background}}
& \multicolumn{5}{c}{\textbf{Edited}}
& \multicolumn{2}{c}{\textbf{Properties}}
& \multicolumn{4}{c}{\textbf{Efficiency}} \\
\cmidrule(lr){3-8}
\cmidrule(lr){9-13}
\cmidrule(lr){14-15}
\cmidrule(lr){16-19}
\multicolumn{1}{c}{} & \multicolumn{1}{c}{} & \multicolumn{1}{c}{\textbf{PSNR}$\uparrow$} & \multicolumn{1}{c}{\textbf{MSE}$\downarrow$} & \multicolumn{1}{c}{\textbf{LPIPS}$\downarrow$} & \multicolumn{1}{c}{\textbf{SSIM}$\uparrow$} & \multicolumn{1}{c}{\textbf{DINO}$\downarrow$} & \multicolumn{1}{c}{\textbf{CLIP}$\uparrow$} & \multicolumn{1}{c}{\textbf{CLIP-W}$\uparrow$} & \multicolumn{1}{c}{\textbf{CLIP-E}$\uparrow$} & \multicolumn{1}{c}{\textbf{CLIP-D}$\uparrow$} & \multicolumn{1}{c}{\textbf{MUSIQ}$\uparrow$} & \multicolumn{1}{c}{\textbf{MUSIQ-E}$\uparrow$} & \multicolumn{1}{c}{\textbf{T-free}} & \multicolumn{1}{c}{\textbf{I-free}} & \multicolumn{1}{c}{\textbf{Step}} & \multicolumn{1}{c}{\textbf{NFE}} & \multicolumn{1}{c}{\textbf{Runtime(s)}$\downarrow$} & \multicolumn{1}{c}{\textbf{VRAM(MB)}$\downarrow$} \\
\midrule
\multirow{4}{*}{\shortstack{Multi-step\\($>$10 steps)}}
& SpotEdit & 27.21 & 13.92 & 69.39 & 0.88 & 41.55 & 96.64 & 24.64 & 22.51 & 17.63 & 67.01 & 60.04 & \gcheck & \gcheck & 50 & 50 & 11.37 & 26384 \\
& FlowEdit & 22.12 & 8.91 & 104.71 & 0.84 & 26.88 & 95.12 & 26.01 & 22.78 & 10.90 & 69.39 & 63.72 & \gcheck & \gcheck & 33 & 33 & 9.39 & 20180 \\
& KV-Edit & 34.91 & 0.52 & 11.26 & 0.95 & 16.57 & 97.06 & 24.56 & 20.83 & 7.95 & 71.11 & 62.10 & \gcheck & \rcross & 24 & 48 & 9.42 & 31618 \\
& PnPInversion & 22.32 & 8.29 & 112.77 & 0.79 & 28.11 & 95.03 & 25.42 & 22.53 & 10.43 & 68.82 & 61.17 & \gcheck & \rcross & 50 & 100 & 15.03 & 9290 \\
\midrule
\multirow{3}{*}{\shortstack{Few-step\\(2--10 steps)}}
& TurboEdit & 22.55 & 9.25 & 101.93 & 0.80 & 29.87 & 95.24 & 26.54 & 23.04 & 14.62 & 71.11 & 63.83 & \gcheck & \gcheck & 4 & 4 & 0.91 & 19574 \\
& InstantEdit & 28.61 & 2.79 & 45.21 & 0.87 & 12.72 & 97.41 & 24.89 & 21.93 & 13.06 & 67.38 & 60.41 & \gcheck & \rcross & 4 & 8 & 0.84 & 7812 \\
& InfEdit & 28.11 & 6.98 & 56.09 & 0.85 & 16.90 & 97.32 & 24.87 & 22.07 & 9.65 & 67.54 & 60.83 & \gcheck & \gcheck & 4 & 4 & 1.29 & 6794 \\
\midrule
\multirow{5}{*}{\shortstack{One-step\\(1 step)}}
& SwiftEdit & \cellcolor{secondfill}24.22 & \cellcolor{bestfill}5.41 & \cellcolor{bestfill}76.39 & \cellcolor{bestfill}0.82 & \cellcolor{bestfill}15.14 & \cellcolor{bestfill}96.34 & 24.48 & 21.54 & 7.56 & 66.52 & 60.05 & \rcross & \rcross & 1 & 2 & 0.45 & 19500 \\
& ChordEdit & 22.63 & 7.98 & 122.66 & 0.77 & 28.81 & 95.04 & 24.83 & 22.15 & 10.72 & \cellcolor{secondfill}67.21 & \cellcolor{secondfill}60.19 & \gcheck & \gcheck & 1 & 2 & \cellcolor{secondfill}0.26 & \cellcolor{secondfill}7333 \\
& WhereEdit & 24.05 & 7.97 & 108.22 & 0.80 & 31.03 & 94.93 & 25.18 & \cellcolor{bestfill}22.66 & 10.78 & 64.28 & 56.39 & \gcheck & \gcheck & 1 & 3 & 0.29 & 7339 \\
& RRLS & 23.92 & 6.27 & 111.40 & \cellcolor{secondfill}0.81 & 25.07 & 95.21 & \cellcolor{bestfill}25.38 & 22.09 & \cellcolor{secondfill}11.47 & 65.90 & 57.79 & \gcheck & \gcheck & 1 & 4 & 0.52 & \cellcolor{secondfill}7333 \\
& \textbf{BudEdit} & \cellcolor{bestfill}\textbf{24.71} & \cellcolor{secondfill}\textbf{5.54} & \cellcolor{secondfill}\textbf{78.24} & \cellcolor{bestfill}\textbf{0.82} & \cellcolor{secondfill}\textbf{20.01} & \cellcolor{secondfill}\textbf{96.17} & \cellcolor{secondfill}\textbf{25.25} & \cellcolor{secondfill}\textbf{22.52} & \cellcolor{bestfill}\textbf{12.28} & \cellcolor{bestfill}\textbf{67.77} & \cellcolor{bestfill}\textbf{60.31} & \gcheck & \gcheck & 1 & 2 & \cellcolor{bestfill}\textbf{0.21} & \cellcolor{bestfill}\textbf{7319} \\
\bottomrule
\end{tabular}
}
\vskip 3pt
\includegraphics[width=\textwidth]{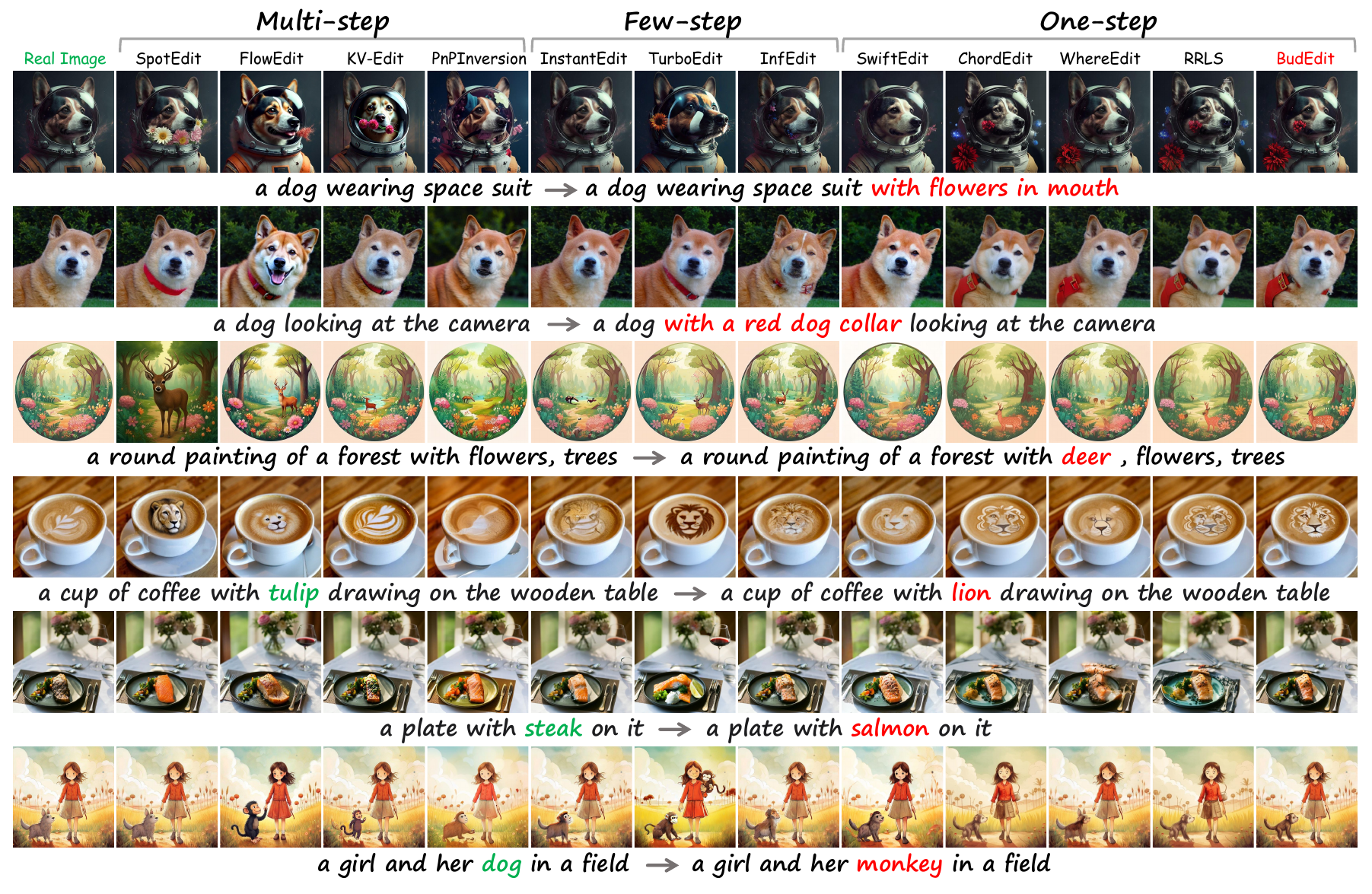}
\vskip -10pt
\captionof{figure}{\textbf{Qualitative comparison with the methods of
Table~\ref{tab:comparison}}. Several baselines exhibit background color shifts, ghosting, or structural
distortion, while BudEdit better preserves the source surroundings and
produces detailed edits with clean boundaries, while preserving the intended edit semantics throughout.}
\label{fig:qualitative}
\end{table}

\paragraph{Metrics.}
Eleven metrics, one per column of Table~\ref{tab:comparison}: background-fidelity (PSNR, MSE, LPIPS~\citep{zhang2018lpips}, SSIM,
DINO~\citep{caron2021dino}, CLIP~\citep{radford2021clip}) and editing-quality
(CLIP-W, CLIP-E, CLIP-D~\citep{gal2022stylegannada}, MUSIQ, MUSIQ-E~\citep{ke2021musiq}).
\paragraph{Implementation.}
In Table~\ref{tab:comparison}, Step counts sampling updates, while the number
of function evaluations (NFE) counts UNet forward evaluations, with the source
and target conditions batched into one evaluation. Like ChordEdit, BudEdit uses one sampling step and two UNet
evaluations, with $t_s=0.78$, $t_e=0.38$, $\alpha=0.7$, $\beta=4.0$;
runtimes are measured on an RTX 4090.
\subsection{Main Results}
\label{sec:main}
\paragraph{Quantitative Analysis.} Table~\ref{tab:comparison} compares BudEdit with existing editors on
PIE-Bench. Among one-step methods, BudEdit achieves the highest PSNR, CLIP-D,
MUSIQ, and MUSIQ-E, while using two UNet evaluations and reporting the lowest
runtime of $0.21$\,s per image.
Among training-free one-step editors, BudEdit leads all six background-fidelity
metrics of Table~\ref{tab:comparison} and exceeds ChordEdit on all five
editing-quality metrics; SwiftEdit reaches its background numbers with
task-specific training. The few-step and multi-step methods provide a broader
quality--compute reference rather than a same-budget ablation: they use
$4$--$100$ evaluations and report $0.84$--$15.03$\,s per image, compared
with BudEdit's two evaluations and $0.21$\,s.

\paragraph{Qualitative Analysis.} Figure~\ref{fig:qualitative} shows the corresponding qualitative comparison.
In the lion latte-art example, BudEdit produces a clearly delineated mane while
preserving the cup and table appearance of the source image; in the
steak-to-salmon example it preserves the dark plate and side dishes. These
examples complement Table~\ref{tab:comparison}: BudEdit combines strong
background fidelity with the highest CLIP-D among one-step methods while
retaining fine target texture and clean edit boundaries.

\begin{wraptable}[19]{r}{0.50\textwidth}
\captionsetup{skip=1pt, belowskip=0pt, font=footnotesize}
\centering
\caption{\textbf{ChordEdit chord window ablation.}
$\delta{=}0.15$ is default; $\delta{=}0$ disables the window.}
\vskip 1pt
\label{tab:controlled}
\scriptsize
\setlength{\tabcolsep}{1.2pt}
\resizebox{\linewidth}{!}{%
\begin{tabular}{lccccc}
\toprule
Config & PSNR$\uparrow$ & MSE$\downarrow$ & CLIP-W$\uparrow$ & CLIP-E$\uparrow$ & MUSIQ-E$\uparrow$ \\
\midrule
$\delta{=}0$ & 20.13 & 13.34 & 25.11 & \textbf{22.54} & 55.97 \\
$\delta{=}0.15$ & 22.63 & 7.98 & 24.83 & 22.15 & 60.19 \\
\rowcolor{secondfill}
\textbf{BudEdit} & \textbf{24.71} & \textbf{5.54} & \textbf{25.25} & 22.52 & \textbf{60.31} \\
\bottomrule
\end{tabular}%
}
\vskip 6pt
\vspace{-4pt}
\includegraphics[width=0.95\linewidth]{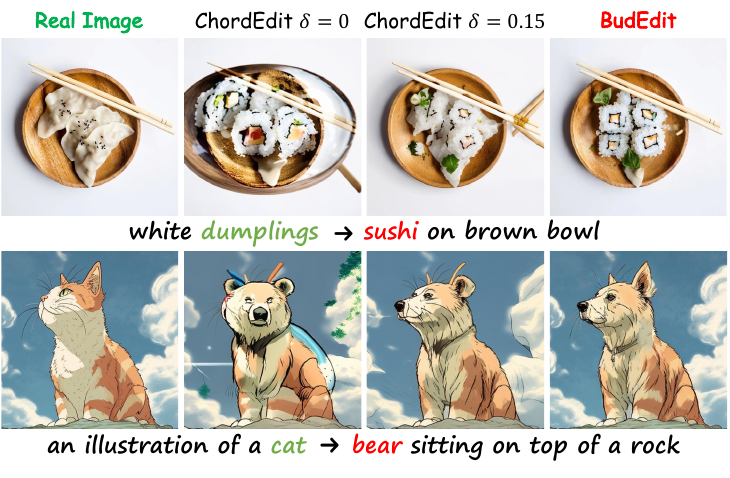}
\vskip -1pt
\captionof{figure}{\textbf{ChordEdit window ablation.} ChordEdit with $\delta{=}0$ distorts the scene; BudEdit preserves scene.}
\label{fig:chord-naive}
\end{wraptable}

\paragraph{Comparison with Chord transport.} In Table~\ref{tab:controlled}, we examine the role of temporal smoothing in ChordEdit. Disabling its chord window ($\delta{=}0$) raises CLIP-E from $22.15$ to $22.54$, but lowers MUSIQ-E from $60.19$ to $55.97$ and background PSNR from $22.63$ to $20.13$\,dB. Figure~\ref{fig:chord-naive} further shows visible background and structural distortions in the window-off outputs. CLIP-E alone therefore does not fully reflect edit quality, and we include MUSIQ-E as a complementary measure of edited-region perceptual quality. Without relying on temporal smoothing, BudEdit improves all five metrics in
Table~\ref{tab:controlled} over ChordEdit's smoothed default. Compared with the window-off row, BudEdit gives up only $0.02$ CLIP-E while improving background PSNR by $4.58$\,dB and MUSIQ-E by $4.34$ points.

\paragraph{Matched effective timestep.} ChordEdit's default chord window assigns six sevenths of its weight to $t_s-\delta$. With $t_s{=}0.9$ and $\delta{=}0.15$, its weighted-average time is $t_s-6\delta/7=0.771$, close to BudEdit's direct timestep $t_s{=}0.78$. This proximity is consistent with temporal smoothing shifting ChordEdit toward a lower effective timestep, while the mechanisms remain different: ChordEdit averages fields globally over time, whereas BudEdit evaluates directly at $t_s{=}0.78$ and controls the trade-off through spatial energy budgeting. This reading is corroborated by an independent reproduction of ChordEdit~\citep{li2026rethinking}.

\begin{figure}[t]
\centering
\includegraphics[width=0.8\textwidth]{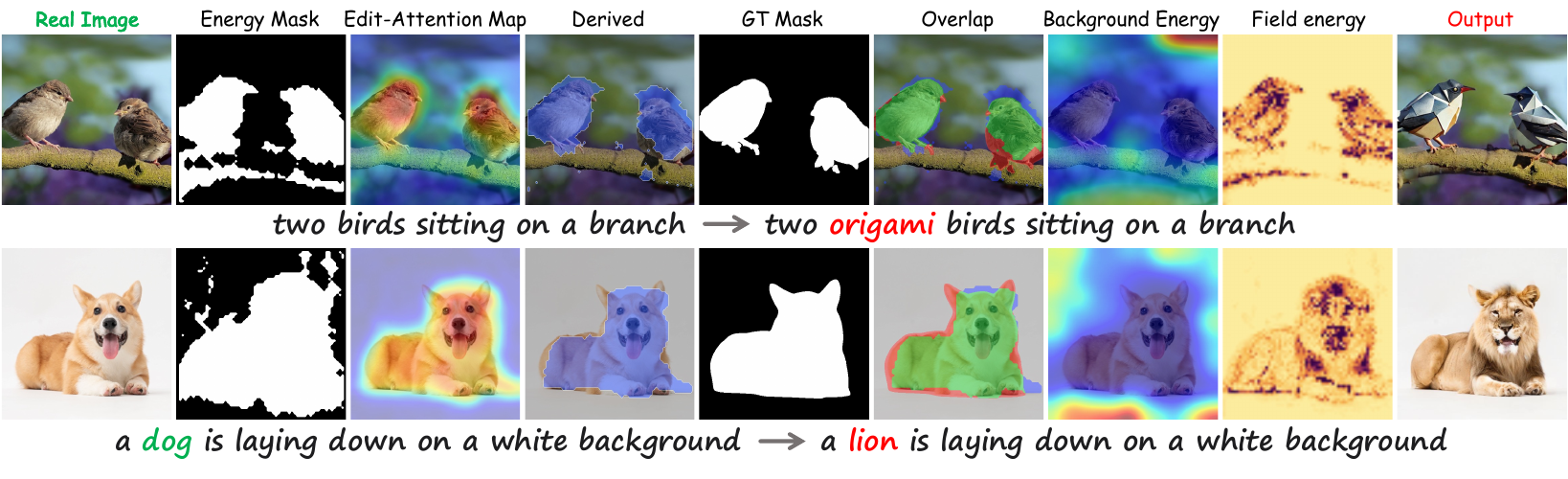}
\captionsetup{skip=1pt, belowskip=0pt}
\vskip -1pt
\caption{\textbf{Annotation-free localization.} Two edits, from birds to origami birds and from a dog to a lion. The columns show the Real Image, Energy Mask, Edit-Attention Map, Derived Mask, GT Mask, Overlap, background energy, field energy of the composed transport field, and output.}
\label{fig:localization}
\end{figure}

\subsection{Verification of the Theory's Quantities}
\label{sec:interpretability}
\paragraph{Localization derives from the model itself.} Figure~\ref{fig:localization} illustrates this model-derived localization on two examples. The GT Mask is included for visual comparison only; editing uses no external or
human-annotated mask, and the Derived Mask is produced by the model itself. In the Overlap panel, green marks the common support of the Derived Mask and the GT Mask, while red and blue mark regions selected by only one of them. The appendix reports the overlap between the derived support and the annotated region over all $700$ pairs.

\vspace{-7pt}
\subsection{Component Ablations}
\label{sec:ablations}

Table~\ref{tab:ablation} isolates budgeted energy reallocation from second-pass cleanup. Setting $\beta{=}0$ drops the
\begin{wraptable}[18]{r}{0.5\textwidth}
\captionsetup{skip=1pt, belowskip=0pt, font=footnotesize}
\centering
\caption{\textbf{Factorial ablation of energy reallocation and cleanup.}Bold: column best.
}
\vskip 1pt
\label{tab:ablation}
\scriptsize
\setlength{\tabcolsep}{1.1pt}
\resizebox{\linewidth}{!}{%
\begin{tabular}{cccccccc}
\toprule
Energy
& Cleanup
& PSNR$\uparrow$
& SSIM$\uparrow$
& CLIP-W$\uparrow$
& CLIP-E$\uparrow$
& CLIP-D$\uparrow$
& NFE$\downarrow$ \\
\midrule
\rcross & \rcross
& \textbf{26.14} & \textbf{0.85} & 23.51 & 20.51 & 3.48 & \textbf{1} \\

\gcheck & \rcross
& 25.73 & 0.84 & 23.73 & 20.82 & 5.05 & \textbf{1} \\

\rcross & \gcheck
& 25.92 & 0.83 & 24.59 & 21.81 & 9.35 & 2 \\

\rowcolor{secondfill}
\gcheck & \gcheck
& 24.71
& 0.82
& \textbf{25.25}
& \textbf{22.52}
& \textbf{12.28}
& 2 \\
\bottomrule
\end{tabular}%
}
\vspace{-4pt}
\includegraphics[width=0.9\linewidth]{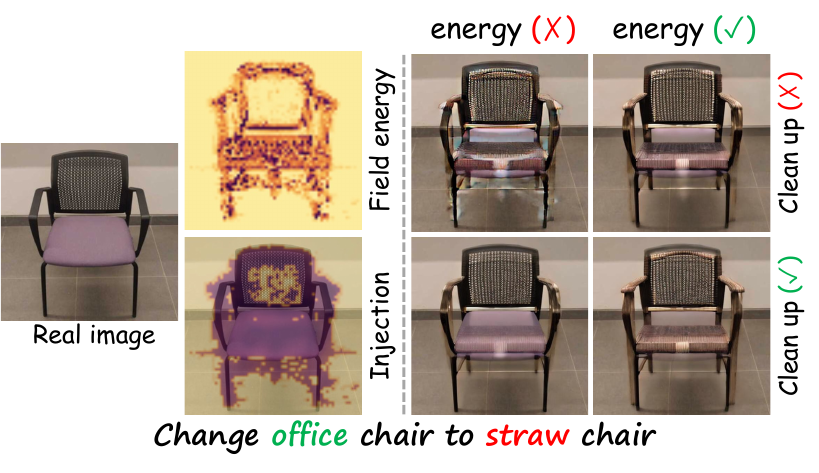}
\captionof{figure}{\textbf{Qualitative and energy analysis.} Injection: energy of $f^\star$; field energy: energy of the transport field.}
\label{fig:ablation-qualitative}
\end{wraptable}
budgeted injection field $f^{\star}$ from Eq.~\ref{eq:field}, leaving only the attenuated residual term; disabling cleanup drops the second UNet evaluation, halving NFE.
Each factor helps alone. Energy reallocation alone raises CLIP-E/CLIP-D by $0.31$/$1.57$ without
cleanup; cleanup gives the larger editing-quality gain, $+1.30$/$+5.87$ and $+1.08$ CLIP-W. They
also interact: with cleanup on, the reallocation gains grow to $0.71$/$2.93$, and the full
configuration attains the best CLIP-W, CLIP-E and CLIP-D. Transporting less energy changes the background less, so the no-energy, no-cleanup setting holds the highest PSNR and SSIM. Cleanup recovers under-editing with an extra forward pass, whereas energy reallocation goes further within a single forward pass: it moves the measured background energy into the injection support, tightening editing strength without an additional evaluation, unlike temporal smoothing.\par
Figure~\ref{fig:ablation-qualitative} visualizes this interaction: the injection energy map shows where $f^{\star}$ spends the budget, and the field energy map shows the resulting field of Eq.~\ref{eq:field}. For the \emph{office chair} $\to$ \emph{straw chair} edit, budgeted energy reallocation alone overlays straw texture on the cushion, cleanup alone yields only a slight change to the armrests, and only the full configuration converts the chair coherently.

\begin{wraptable}{r}{0.5\textwidth}
\captionsetup{font=footnotesize}
\centering
\caption{\textbf{Quantitative comparison on different datasets and backbones.} Bold: column best.}
\vskip 2pt
\vspace{-10pt}
\label{tab:cross_generalization}
\footnotesize
\renewcommand{\arraystretch}{1.0}
\setlength{\tabcolsep}{1.5pt}
\resizebox{\linewidth}{!}{%
\begin{tabular}{llc>{\columncolor{secondfill}}c c>{\columncolor{secondfill}}c}
\toprule
\multirow[c]{2}{*}{Dataset} & \multirow[c]{2}{*}{Backbone} & \multicolumn{2}{c}{PSNR$\uparrow$} & \multicolumn{2}{c}{CLIP-E$\uparrow$} \\
\cmidrule(lr){3-4}\cmidrule(lr){5-6}
& & ChordEdit & \cellcolor{secondfill}\textbf{BudEdit} & ChordEdit & \cellcolor{secondfill}\textbf{BudEdit} \\
\midrule
\multirow[c]{2}{*}{PIE-Bench}
& SD-Turbo & 22.63 & \textbf{24.71} & 22.15 & \textbf{22.52} \\
& SwiftBrushV2 & 22.93 & \textbf{25.49} & 21.18 & \textbf{21.26} \\
\midrule
\multirow[c]{2}{*}{PIE-Bench++}
& SD-Turbo & 22.87 & \textbf{25.06} & 24.12 & \textbf{24.27} \\
& SwiftBrushV2 & 23.16 & \textbf{25.83} & 22.46 & \textbf{22.70} \\
\bottomrule
\end{tabular}%
}
\end{wraptable}

\vspace{-2pt}
\subsection{Cross-Dataset and Cross-Backbone Analysis}
Table~\ref{tab:cross_generalization} crosses two datasets,
PIE-Bench~\citep{ju2024pnp} and the multi-aspect PIE-Bench++ of
ParallelEdits~\citep{huang2024paralleledits}, with
two one-step backbones, SD-Turbo~\citep{sauer2023adversarial}
and SwiftBrushV2~\citep{nguyen2024swiftbrushv2}, using the same hyperparameters
throughout. BudEdit achieves the best background preservation and editing
quality across all four settings, improving background PSNR by over $2$\,dB.
The appendix reports all eleven metrics.

\vspace{-3pt}
\section{Conclusion}
\label{sec:conclusion}

BudEdit addresses spatially misallocated updates in one-step diffusion editing by converting single-draw background residual energy into an image-dependent budget and re-injecting it into the regions selected jointly by residual energy and cross-attention, with exact injection-budget matching and zero injection on the operational background support. On PIE-Bench, it outperforms the temporal-smoothing approach of ChordEdit on all 11 reported metrics, including a $2.1$\,dB PSNR gain, at the lowest runtime in the comparison. Together, these results establish explicit spatial energy budgeting as an alternative to temporal smoothing for stable, efficient, controllable one-step editing.

\bibliography{iclr2027_conference}

@inproceedings{song2020denoising,
  title     = {Denoising Diffusion Implicit Models},
  author    = {Song, Jiaming and Meng, Chenlin and Ermon, Stefano},
  booktitle = {International Conference on Learning Representations (ICLR)},
  year      = {2021}
}

@inproceedings{song2021score,
  title     = {Score-Based Generative Modeling through Stochastic Differential Equations},
  author    = {Song, Yang and Sohl-Dickstein, Jascha and Kingma, Diederik P. and Kumar, Abhishek and Ermon, Stefano and Poole, Ben},
  booktitle = {International Conference on Learning Representations (ICLR)},
  year      = {2021}
}

@inproceedings{salimans2022distillation,
  title     = {Progressive Distillation for Fast Sampling of Diffusion Models},
  author    = {Salimans, Tim and Ho, Jonathan},
  booktitle = {International Conference on Learning Representations (ICLR)},
  year      = {2022}
}

@inproceedings{sauer2023adversarial,
  title     = {Adversarial Diffusion Distillation},
  author    = {Sauer, Axel and Lorenz, Dominik and Blattmann, Andreas and Rombach, Robin},
  booktitle = {Proceedings of the European Conference on Computer Vision (ECCV)},
  pages     = {87--103},
  year      = {2024}
}

@inproceedings{song2023consistency,
  title     = {Consistency Models},
  author    = {Song, Yang and Dhariwal, Prafulla and Chen, Mark and Sutskever, Ilya},
  booktitle = {Proceedings of the 40th International Conference on Machine Learning},
  series    = {Proceedings of Machine Learning Research},
  volume    = {202},
  pages     = {32211--32252},
  publisher = {PMLR},
  year      = {2023}
}

@inproceedings{liu2023instaflow,
  title     = {{InstaFlow}: One Step is Enough for High-Quality Diffusion-Based Text-to-Image Generation},
  author    = {Liu, Xingchao and Zhang, Xiwen and Ma, Jianzhu and Peng, Jian and Liu, Qiang},
  booktitle = {International Conference on Learning Representations (ICLR)},
  year      = {2024}
}

@inproceedings{nguyen2023swiftbrush,
  title     = {{SwiftBrush}: One-Step Text-to-Image Diffusion Model with Variational Score Distillation},
  author    = {Nguyen, Thuan Hoang and Tran, Anh},
  booktitle = {Proceedings of the IEEE/CVF Conference on Computer Vision and Pattern Recognition (CVPR)},
  pages     = {7807--7816},
  year      = {2024}
}

@inproceedings{nguyen2024swiftbrushv2,
  title     = {{SwiftBrush} v2: Make Your One-step Diffusion Model Better Than Its Teacher},
  author    = {Dao, Trung and Nguyen, Thuan Hoang and Le, Thanh and Vu, Duc and Nguyen, Khoi and Pham, Cuong and Tran, Anh},
  booktitle = {Proceedings of the European Conference on Computer Vision (ECCV)},
  pages     = {176--192},
  year      = {2024}
}

@inproceedings{hertz2022prompt,
  title     = {Prompt-to-Prompt Image Editing with Cross-Attention Control},
  author    = {Hertz, Amir and Mokady, Ron and Tenenbaum, Jay and Aberman, Kfir and Pritch, Yael and Cohen-Or, Daniel},
  booktitle = {International Conference on Learning Representations (ICLR)},
  year      = {2023}
}

@inproceedings{tumanyan2023plug,
  title     = {Plug-and-Play Diffusion Features for Text-Driven Image-to-Image Translation},
  author    = {Tumanyan, Narek and Geyer, Michal and Bagon, Shai and Dekel, Tali},
  booktitle = {Proceedings of the IEEE/CVF Conference on Computer Vision and Pattern Recognition (CVPR)},
  pages     = {1921--1930},
  year      = {2023}
}

@inproceedings{brooks2023instructpix2pix,
  title     = {{InstructPix2Pix}: Learning to Follow Image Editing Instructions},
  author    = {Brooks, Tim and Holynski, Aleksander and Efros, Alexei A.},
  booktitle = {Proceedings of the IEEE/CVF Conference on Computer Vision and Pattern Recognition (CVPR)},
  pages     = {18392--18402},
  year      = {2023}
}

@inproceedings{mokady2023nulltext,
  title     = {NULL-Text Inversion for Editing Real Images Using Guided Diffusion Models},
  author    = {Mokady, Ron and Hertz, Amir and Aberman, Kfir and Pritch, Yael and Cohen-Or, Daniel},
  booktitle = {Proceedings of the IEEE/CVF Conference on Computer Vision and Pattern Recognition (CVPR)},
  pages     = {6038--6047},
  year      = {2023}
}

@inproceedings{kulikov2025flowedit,
  title     = {{FlowEdit}: Inversion-Free Text-Based Editing Using Pre-Trained Flow Models},
  author    = {Kulikov, Vladimir and Kleiner, Matan and Huberman-Spiegelglas, Inbar and Michaeli, Tomer},
  booktitle = {Proceedings of the IEEE/CVF International Conference on Computer Vision (ICCV)},
  pages     = {19721--19730},
  year      = {2025}
}

@inproceedings{xu2024infedit,
  title     = {Inversion-Free Image Editing with Language-Guided Diffusion Models},
  author    = {Xu, Sihan and Huang, Yidong and Pan, Jiayi and Ma, Ziqiao and Chai, Joyce},
  booktitle = {Proceedings of the IEEE/CVF Conference on Computer Vision and Pattern Recognition (CVPR)},
  pages     = {9452--9461},
  year      = {2024}
}

@inproceedings{nguyen2025swiftedit,
  title     = {{SwiftEdit}: Lightning Fast Text-Guided Image Editing via One-Step Diffusion},
  author    = {Nguyen, Trong-Tung and Nguyen, Quang and Nguyen, Khoi and Tran, Anh and Pham, Cuong},
  booktitle = {Proceedings of the IEEE/CVF Conference on Computer Vision and Pattern Recognition (CVPR)},
  pages     = {21492--21501},
  year      = {2025}
}

@inproceedings{lu2026chordedit,
  title     = {{ChordEdit}: One-Step Low-Energy Transport for Image Editing},
  author    = {Lu, Liangsi and Chen, Xuhang and Guo, Minzhe and Li, Shichu and Wang, Jingchao and Shi, Yang},
  booktitle = {Proceedings of the IEEE/CVF Conference on Computer Vision and Pattern Recognition (CVPR)},
  pages     = {14398--14407},
  year      = {2026}
}

@inproceedings{zhu2025kvedit,
  title     = {{KV-Edit}: Training-Free Image Editing for Precise Background Preservation},
  author    = {Zhu, Tianrui and Zhang, Shiyi and Shao, Jiawei and Tang, Yansong},
  booktitle = {Proceedings of the IEEE/CVF International Conference on Computer Vision (ICCV)},
  pages     = {16607--16617},
  year      = {2025}
}

@inproceedings{qin2025spotedit,
  title     = {{SpotEdit}: Selective Region Editing in Diffusion Transformers},
  author    = {Qin, Zhibin and Tan, Zhenxiong and Wang, Zeqing and Liu, Songhua and Wang, Xinchao},
  booktitle = {Proceedings of the IEEE/CVF Conference on Computer Vision and Pattern Recognition (CVPR)},
  pages     = {18683--18692},
  year      = {2026}
}

@inproceedings{jiang2025flowdc,
  title     = {{FlowDC}: Flow-Based Decoupling-Decay for Complex Image Editing},
  author    = {Jiang, Yilei and Wang, Zhen and Wang, Yanghao and Yu, Jun and Zhuang, Yueting and Xiao, Jun and Chen, Long},
  booktitle = {Proceedings of the IEEE/CVF Conference on Computer Vision and Pattern Recognition (CVPR)},
  pages     = {25757--25766},
  year      = {2026}
}

@article{whereedit,
  title     = {{WhereEdit}: Mask-aware Local Latent Editing for One-Step Image Editing},
  author    = {Hu, Ming and Dou, Mingyu and Yin, Jianfu and Zhang, Miaomiao and Hu, Cong and Wang, Yao and Hu, Bingliang and Wang, Quan},
  journal   = {arXiv preprint arXiv:2607.20883},
  year      = {2026}
}

@article{rrls,
  title     = {Bridging the Manifold Gap: Riemannian Residual Line Search for One-Step Image Editing},
  author    = {Yi, Hongzhu and Luo, Zhongtian and Li, Tong and Fan, Yiyan and Xu, Jungang},
  journal   = {arXiv preprint arXiv:2606.24844},
  year      = {2026}
}

@inproceedings{ju2024pnp,
  title     = {{PnP} Inversion: Boosting Diffusion-based Editing with 3 Lines of Code},
  author    = {Ju, Xuan and Zeng, Ailing and Bian, Yuxuan and Liu, Shaoteng and Xu, Qiang},
  booktitle = {International Conference on Learning Representations (ICLR)},
  year      = {2024}
}

@article{benamou2000computational,
  title     = {A Computational Fluid Mechanics Solution to the {Monge-Kantorovich} Mass Transfer Problem},
  author    = {Benamou, Jean-David and Brenier, Yann},
  journal   = {Numerische Mathematik},
  volume    = {84},
  number    = {3},
  pages     = {375--393},
  year      = {2000},
  doi       = {10.1007/s002110050002}
}

@inproceedings{lipman2022flowmatching,
  title     = {Flow Matching for Generative Modeling},
  author    = {Lipman, Yaron and Chen, Ricky T. Q. and Ben-Hamu, Heli and Nickel, Maximilian and Le, Matt},
  booktitle = {International Conference on Learning Representations (ICLR)},
  year      = {2023}
}

@article{chefer2023attend,
  title     = {{Attend-and-Excite}: Attention-Based Semantic Guidance for Text-to-Image Diffusion Models},
  author    = {Chefer, Hila and Alaluf, Yuval and Vinker, Yael and Wolf, Lior and Cohen-Or, Daniel},
  journal   = {ACM Transactions on Graphics},
  volume    = {42},
  number    = {4},
  articleno = {148},
  numpages  = {10},
  year      = {2023},
  doi       = {10.1145/3592116}
}

@inproceedings{caron2021dino,
  title     = {Emerging Properties in Self-Supervised Vision Transformers},
  author    = {Caron, Mathilde and Touvron, Hugo and Misra, Ishan and J{\'e}gou, Herv{\'e} and Mairal, Julien and Bojanowski, Piotr and Joulin, Armand},
  booktitle = {Proceedings of the IEEE/CVF International Conference on Computer Vision (ICCV)},
  pages     = {9650--9660},
  year      = {2021}
}

@inproceedings{deutch2024turboedit,
  title     = {{TurboEdit}: Text-Based Image Editing Using Few-Step Diffusion Models},
  author    = {Deutch, Gilad and Gal, Rinon and Garibi, Daniel and Patashnik, Or and Cohen-Or, Daniel},
  booktitle = {SIGGRAPH Asia 2024 Conference Papers},
  pages     = {1--12},
  publisher = {ACM},
  year      = {2024},
  doi       = {10.1145/3680528.3687612}
}

@inproceedings{instantedit,
  title     = {{InstantEdit}: Text-Guided Few-Step Image Editing with Piecewise Rectified Flow},
  author    = {Gong, Yiming and Zhu, Zhen and Zhang, Minjia},
  booktitle = {Proceedings of the IEEE/CVF International Conference on Computer Vision (ICCV)},
  pages     = {16808--16817},
  year      = {2025}
}

@article{li2026rethinking,
  title   = {Rethinking One-Step Image Editing through ChordEdit: Reproduction, Simplification, and New Insights},
  author  = {Li, Minghan and Moebel, Jeremy and Wang, Mengyu},
  journal = {arXiv preprint arXiv:2606.14042},
  year    = {2026}
}

@inproceedings{ke2021musiq,
  title     = {{MUSIQ}: Multi-Scale Image Quality Transformer},
  author    = {Ke, Junjie and Wang, Qifei and Wang, Yilin and Milanfar, Peyman and Yang, Feng},
  booktitle = {Proceedings of the IEEE/CVF International Conference on Computer Vision (ICCV)},
  pages     = {5148--5157},
  year      = {2021}
}

@inproceedings{huang2024paralleledits,
  title     = {{ParallelEdits}: Efficient Multi-Aspect Text-Driven Image Editing with Attention Grouping},
  author    = {Huang, Mingzhen and Cai, Jialing and Jia, Shan and Lokhande, Vishnu Suresh and Lyu, Siwei},
  booktitle = {Advances in Neural Information Processing Systems (NeurIPS)},
  volume    = {37},
  pages     = {22569--22595},
  year      = {2024},
  doi       = {10.52202/079017-0710}
}

@article{gal2022stylegannada,
  title   = {{StyleGAN-NADA}: {CLIP}-Guided Domain Adaptation of Image Generators},
  author  = {Gal, Rinon and Patashnik, Or and Maron, Haggai and Bermano, Amit H. and Chechik, Gal and Cohen-Or, Daniel},
  journal = {ACM Transactions on Graphics},
  volume  = {41},
  number  = {4},
  pages   = {1--13},
  month   = {jul},
  year    = {2022},
  doi     = {10.1145/3528223.3530164}
}

@inproceedings{radford2021clip,
  title     = {Learning Transferable Visual Models From Natural Language Supervision},
  author    = {Radford, Alec and Kim, Jong Wook and Hallacy, Chris and Ramesh, Aditya and Goh, Gabriel and Agarwal, Sandhini and Sastry, Girish and Askell, Amanda and Mishkin, Pamela and Clark, Jack and Krueger, Gretchen and Sutskever, Ilya},
  booktitle = {Proceedings of the 38th International Conference on Machine Learning (ICML)},
  pages     = {8748--8763},
  year      = {2021}
}

@inproceedings{zhang2018lpips,
  title     = {The Unreasonable Effectiveness of Deep Features as a Perceptual Metric},
  author    = {Zhang, Richard and Isola, Phillip and Efros, Alexei A. and Shechtman, Eli and Wang, Oliver},
  booktitle = {Proceedings of the IEEE Conference on Computer Vision and Pattern Recognition (CVPR)},
  pages     = {586--595},
  year      = {2018}
}
\bibliographystyle{iclr2027_conference}

\end{document}